%% file: egpaper_final.tex
\documentclass[10pt,twocolumn,letterpaper]{article}

\usepackage{iccv}
\usepackage{times}
\usepackage{epsfig}
\usepackage{graphicx}
\usepackage{amsmath}
\usepackage{amssymb}
\input{math_commands.tex}

\usepackage{booktabs}
\usepackage{multirow}
\usepackage{bm}
\usepackage[breaklinks=true,bookmarks=false]{hyperref}

\usepackage{enumitem}

\iccvfinalcopy % *** Uncomment this line for the final submission

\ificcvfinal\fi
\def\@fnsymbol#1{\ensuremath{\ifcase#1\or \dagger\or \ddagger\or
   \mathsection\or \mathparagraph\or \|\or **\or \dagger\dagger
   \or \ddagger\ddagger \else\@ctrerr\fi}}
\begin{document}

%%%%%%%%% TITLE
\title{Lossy and Lossless (L$^2$) Post-training Model Size Compression}

\author{Yumeng Shi$^{1,2}$ \ \ \ \ \ 
Shihao Bai$^2$ \ \ \ \ \ 
Xiuying Wei$^{1,2}$ \ \ \ \ \ 
Ruihao Gong$^{1,2}$\thanks{Corresponding Author} \ \ \ \ \ 
Jianlei Yang$^{1*}$\\
$^{1}$Beihang University \ \ \ \ \ $^{2}$SenseTime Research\\
{\tt\small \{senbei,jianlei\}@buaa.edu.cn \{baishihao,gongruihao\}@sensetime.com weixiuying966@gmail.com}
}

\maketitle
\ificcvfinal\thispagestyle{empty}\fi

%%%%%%%%% ABSTRACT
\begin{abstract}
Deep neural networks have delivered remarkable performance and have been widely used in various visual tasks. However, their huge size causes significant inconvenience for transmission and storage. Many previous studies have explored model size compression. However, these studies often approach various lossy and lossless compression methods in isolation, leading to challenges in achieving high compression ratios efficiently. This work proposes a post-training model size compression method that combines lossy and lossless compression in a unified way. We first propose a unified parametric weight transformation, which ensures different lossy compression methods can be performed jointly in a post-training manner. Then, a dedicated differentiable counter is introduced to guide the optimization of lossy compression to arrive at a more suitable point for later lossless compression. Additionally, our method can easily control a desired global compression ratio and allocate adaptive ratios for different layers. Finally, our method can achieve a stable $10\times$ compression ratio without sacrificing accuracy and a $20\times$ compression ratio with minor accuracy loss in a short time. Our code is available at \href{https://github.com/ModelTC/L2_Compression}{https://github.com/ModelTC/L2\_Compression}.
\end{abstract}

%%%%%%%%% BODY TEXT
\section{Introduction}

In recent years, deep neural networks (DNNs), especially convolutional neural networks (CNNs) \cite{DBLP:conf/cvpr/HeZRS16, DBLP:conf/cvpr/RadosavovicKGHD20, DBLP:conf/cvpr/SandlerHZZC18, DBLP:conf/cvpr/TanCPVSHL19}, have achieved attractive performance in various computer vision tasks such as image classification, detection, and segmentation. However, as their performance improves, their parameter counts also significantly increase, which is very storage-consuming. Therefore, despite their excellent performance, it is difficult to deploy models with a large number of parameters, particularly on mobile or edge devices with limited storage resources.

Model compression \cite{DBLP:conf/iclr/OktayBSS20, DBLP:journals/corr/HanMD15, DBLP:conf/cvpr/TungM18, DBLP:journals/corr/abs-2102-07725, DBLP:journals/corr/abs-2206-00820, DBLP:conf/iclr/StockFGGGJJ21} is a common solution to reduce the model size, including lossless and lossy compression. Common lossless compression methods such as Huffman coding and Range coding are both entropy coding methods. They can leverage redundant information in data to achieve distortion-free compression. However, for data with weak spatiotemporal coherence, high compression ratios are often difficult to achieve. Compared with lossless methods, lossy methods such as pruning \cite{DBLP:conf/iccvw/LazarevichKM21} and quantization \cite{DBLP:conf/icml/NagelABLB20, DBLP:conf/iccv/NagelBBW19} have attracted more attention recently. Pruning reduces the model size by removing extraneous weights, and quantization replaces weights in a low-bit format. Both pruning and quantization are trade-offs between model distortion and compression ratio. Previous studies have primarily focused on individual compression methods or have merely combined different compression techniques without considering the interaction between them, resulting in multiple isolated trade-offs in successive stages. Hence, they can hardly achieve a higher compression ratio with a small amount of data and little training time.

\begin{figure}[t]
\begin{center}
\includegraphics[width=1\linewidth]{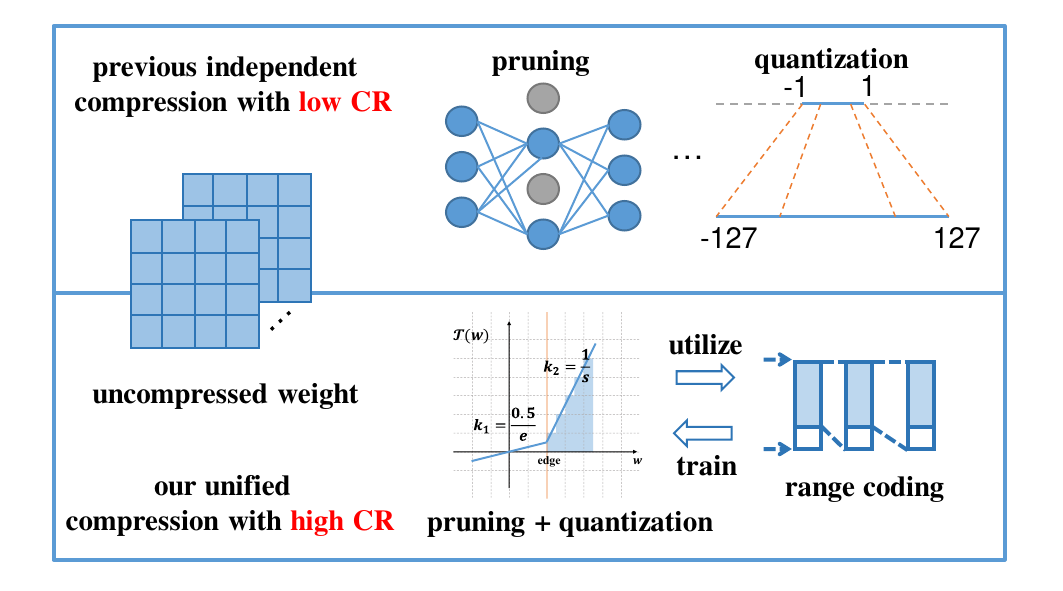}
\end{center}
   \caption{Comparison between previous compression methods and our unified post-training compression.}
\label{fig:intro}
\end{figure}

To address this issue, this paper proposes to mix different compression methods and optimize them under the recent popular post-training setting, which slightly adjusts weights for better performance. We build an optimization objective that introduces an entropy regularization term, making the global compression ratio controllable. Based on it, a unified parametric weight transformation is first designed to integrate different lossy compression techniques together, allowing us to jointly explore various compression strategies and determine unique ones for each layer. Second, we devise a novel differentiable counter to make our entropy regularization term differentiable by leveraging kernel functions. This differentiable way imposes constraints on the distribution of compressed weights during optimization, contributing to more compatible optimized weights with later lossless compression. Consequently, our method combining lossy and lossless compression can achieve a consistently superior compression ratio with satisfying accuracy performance in an efficient manner.

To the best of our knowledge, this is the first work that presents a unified modeling approach for various lossy compression methods while leveraging the characteristics of lossless compression to optimize the process of lossy compression. Extensive experiments on various networks verify the efficacy of our method (e.g., stable $10\times$ compression ratio without sacrificing accuracy and up to $20\times$ compression ratio
with minor accuracy loss).

Our main contributions can be summarized as follows:
\begin{itemize}
\item We propose a pioneering post-training model size compression method that combines lossy and lossless compression with a new optimization objective.
\item We design a unified parametric weight transformation approach for lossy compression methods, integrating techniques such as pruning and quantization into a single stage and determining each layer’s unique compression scheme.
\item We introduce a dedicated differentiable counter to estimate the entropy of compressed weights. This counter ensures that the distribution of the optimized weights is more amenable for lossless compression.
\item Extensive experiments conducted on various architectures, including classification and object detection tasks, demonstrate that our method achieves high compression ratios with negligible accuracy drops.
\end{itemize}

\begin{figure*}[t]
\begin{center}
\includegraphics[width=1\linewidth]{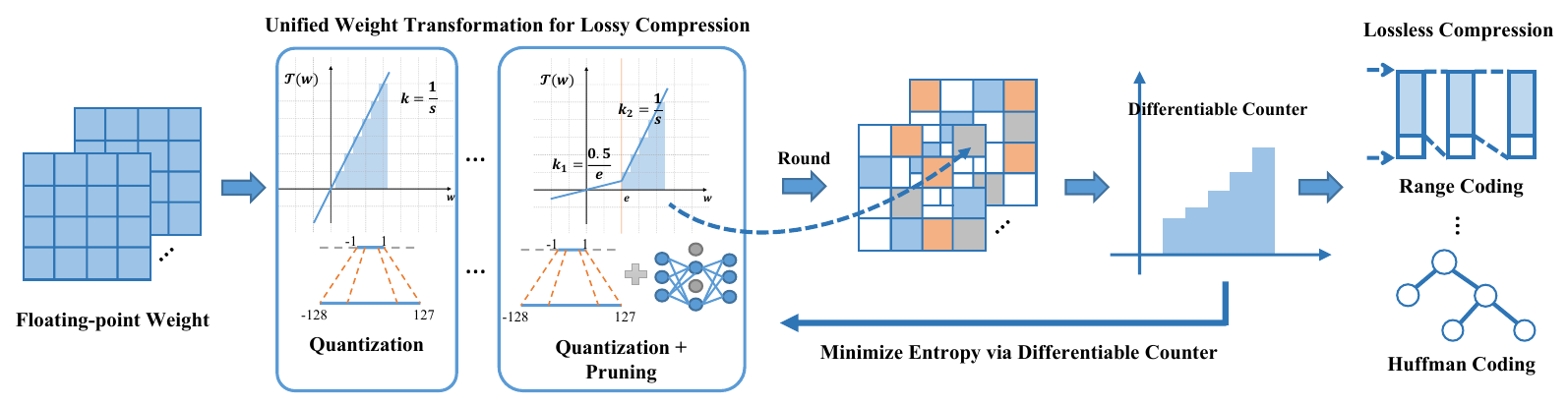}
\end{center}
   \caption{ The Overview of the proposed post-training compression method.}
\label{fig:overview}
\end{figure*}

%-------------------------------------------------------------------------
\section{Related Work}

In the past, numerous researchers have explored model compression techniques. Knowledge distillation \cite{DBLP:journals/corr/HintonVD15} is one such technique that aims to transfer knowledge from a complex teacher model to a simplified student model. By utilizing the soft target probability distribution from the teacher model, the student model achieves comparable performance with a smaller size. Matrix factorization \cite{DBLP:conf/nips/LiebenweinMFR21, DBLP:journals/corr/abs-2207-11048} is another such technique that breaks down neural network weight matrices into smaller ones, reducing parameters and making the model more resource-efficient. Common methods include SVD and QR decomposition. Hereafter, we focus on introducing network pruning, weight quantization, and entropy encoding.

\vspace{1em} \noindent \textbf{Network pruning.} 
This technique achieves model compression by selectively removing unimportant weights. Network pruning can be categorized into structured pruning and unstructured pruning. Structured pruning \cite{DBLP:conf/cvpr/HouQSMYXC00K22} targets specific structures, like rows, columns, channels, or filters, which can accelerate computation. Unstructured pruning \cite{DBLP:conf/iccvw/LazarevichKM21} solely considers the importance of weight elements, disregarding their position. Our objective is to compress the model size, so we focus on unstructured pruning. Previous studies \cite{DBLP:journals/corr/HanPTD15, DBLP:conf/wacv/YuYGDKMK22} have explored various methods to measure weight importance, including magnitude-based and derivative-based methods. Additionally, some studies \cite{DBLP:conf/eccv/HeLLWLH18, DBLP:conf/iccv/YuM021, DBLP:conf/icml/YuM022} have concentrated on determining the sparsity ratio for each layer.

\vspace{1em} \noindent \textbf{Weight quantization.} 
Another common model compression technique is weight quantization, which reduces the number of bits needed for weight storage by discretizing weight values. Many studies \cite{DBLP:conf/icml/NagelABLB20, DBLP:conf/eccv/ZhangYYH18} have explored the effects of quantization intervals, clips, and rounding methods. There are also studies \cite{DBLP:conf/iclr/UllrichMW17, DBLP:conf/cvpr/ParkAY17} focusing on non-uniform quantization, which utilizes clustering to group weights instead of using a uniform formula for calculation. Other studies \cite{DBLP:journals/corr/abs-2302-05397, DBLP:conf/aaai/ZhouMCF18, DBLP:conf/icml/YaoDZGYTW0WMK21, DBLP:conf/iccv/DongYGMK19, DBLP:conf/nips/DongYAGMK20, DBLP:journals/corr/abs-2212-02770, DBLP:journals/corr/abs-2109-07865} have investigated mixed precision quantization, suggesting that the importance of network layers can influence the required storage bit-width for each layer.

\vspace{1em} \noindent \textbf{Entropy coding.} 
Entropy coding, as a lossless data compression technique, aims to achieve data compression without any loss of information. It re-encodes data based on its probability distribution, using shorter codes for high-frequency data and longer codes for low-frequency data, effectively saving storage space. Common entropy coding methods encompass Huffman coding, and arithmetic coding, among others. It is frequently integrated with other methods to further reduce weight storage \cite{DBLP:journals/corr/HanMD15, DBLP:conf/iclr/ChoiEL17} or reduce the memory footprint and transmission bandwidth of feature maps during inference \cite{DBLP:journals/jmlr/BaskinCZBBM21, DBLP:conf/ijcnn/ChmielBZBYKBM20}.
\vspace{1em}

Regarding model compression, using a single technique often fails to achieve satisfactory performance. Certain studies \cite{DBLP:journals/corr/abs-1905-08318, DBLP:journals/jstsp/WiedemannKMHMMN20, DBLP:journals/corr/HanMD15} have explored using multiple methods sequentially. However, they treated them in isolation without considering their mutual impacts. Some studies \cite{DBLP:conf/cvpr/TungM18} have attempted to apply in-parallel pruning and quantization but overlooked the influence of lossless compression on the lossy compression. Furthermore, these approaches typically required extensive training to achieve desirable results. Addressing these challenges involves contemplating the integration of lossless compression effects and the unification of modeling for lossy compression methods. We propose an efficient post-training approach that achieves high compression ratios while maintaining superior accuracies.

%-------------------------------------------------------------------------
\section{Preliminaries}

In this section, some basic concepts of model compression methods will be introduced. Model compression reduces the model size, which can be divided into lossy compression and lossless compression.

\vspace{1em}
\noindent \textbf{Basic notation.} 
In this paper, we mark vectors and flattened matrices $\vw$. Scalar multiplication is defined as $\cdot$. We take $CR$ as an abbreviation for compression ratio, $\mathcal{L}(\cdot)$ as the loss function, $\mathcal{P}(\cdot)$ as probability mass function, and $\mathcal{T}(\cdot)$ as an element-wise transformation that converts dense or real signals to sparse or discrete ones.

\vspace{1em}
\noindent {\textbf{Lossless compression.}
Lossless compression utilizes redundancy in data to achieve compression without introducing distortion, allowing for complete data recovery during decompression. The compression ratio is positively related to the amount of data redundancy. In information theory, Shannon's source coding theorem states that information entropy measures the average information content of given data, representing the average shortest coding length.

Since the model weights do not exhibit spatial and temporal locality in our task, we treat the $m$ elements of each layer's weights $\vw$ as independent and identically distributed random variables $w$. The self-information of $w$ can be expressed as:
\begin{equation}\label{eq1}\mathcal{I}(w)=-log_b\mathcal{P}(w),\end{equation}
where $\mathcal{P(\cdot)}$ represents the probability mass function of $w$. When $b$ equals 2, the unit of self-information is bits. The entropy of $w$ is the expected value of the self-information:
\begin{equation}\label{eq2}\mathcal{H}(w)=\E(\mathcal{I}(w))=-\sum_{\tilde{w}\in\tilde{\vw}}\mathcal{P}(\tilde{w})\log_2\mathcal{P}(\tilde{w}),\end{equation}
where $\tilde{\vw}$ represents the symbol set of $\vw$. Considering $\mathcal{P(\cdot)}$ as the sampling distribution function, the total shortest coding length for a layer is:
\begin{align} \label{eq3}
    \mathcal{S}(\vw)&=m\mathcal{H}(w)=-\sum_{\tilde{w}\in\tilde{\vw}}m\mathcal{P}(\tilde{w})\log_2\mathcal{P}(\tilde{w}) \\ 
    &=-\sum_{\tilde{w}\in\tilde{\vw}}num(\tilde{w})\log_2\mathcal{P}(\tilde{w}) \\
    &=-\sum_{w\in\vw}\log_2\mathcal{P}(w),
\end{align}
where $num(\tilde w)$ represents the number of elements in $\vw$ that are equal to $\tilde w$. The entire neural network's shortest encoding length is the sum of that of each layer.

\vspace{1em}
\noindent \textbf{Lossy compression.}
Pruning and quantization are prevalent lossy compression methods that compress the original weights $\vw$ to $\hat{\vw}$ by removing unimportant parameters or converting floating-point values to low-bit fixed-point numbers. In the post-training setting, some studies~\cite{wei2022qdrop, DBLP:conf/icml/NagelABLB20} on pruning and quantization also utilize a small amount of calibration data to quickly fine-tune the weights. Most studies aim to minimize the mean squared error between the model outputs before and after compression, formulating their optimization goal as follows:
\begin{equation}\label{eq4}
    \min_{\hat{\vw}} \E \left[
    \|\mathcal{F} (\vx, \hat{\vw})) - \mathcal{F} (\vx, \vw)\|_{F}^{2}
    \right],\\
\end{equation}
where $\vx$ is extracted from a calibration dataset with about hundreds of images and $\mathcal{F}(\cdot)$ produces outputs. Here, we use the output of a neural network.

%-------------------------------------------------------------------------
\section{Method}

In this section, we will first present the optimization objective for both lossy and lossless compression methods in the post-training setting. Subsequently, two novel techniques will be introduced: the unified weight transformation and the differentiable counter. These techniques serve to optimize lossy compression uniformly and enhance collaboration with later lossless compression. Using our techniques, it is now possible to efficiently achieve a highly compressed yet accurate model.

\subsection{Optimization Objective}
In this part, a novel optimization objective is introduced, targeting superior compressed results in the post-training setting.

To pursue high accuracy with limited data and GPU effort, we follow previous studies \cite{DBLP:conf/iccvw/LazarevichKM21, DBLP:conf/icml/NagelABLB20} to slightly tune the weights, minimizing the distance between the output after compression and its original counterpart in \autoref{eq4}. Besides, another regularization term is added to encourage small compressed models.
\begin{equation}\label{eq5}
   \min_{\hat{\vw}} \E \left[
    \|\mathcal{F} (\vx, \hat{\vw})) - \mathcal{F} (\vx, \vw)\|_{F}^{2}
     + \lambda \cdot \mathcal{L}_r(\hat{\vw})\right],
\end{equation}
where $\mathcal{L}_r(\cdot)$ defines the new regularization term, and $\lambda$ is the balance factor. 

Motivated by \cite{DBLP:conf/iclr/OktayBSS20}, we incorporate entropy into the regularization term and take $\mathcal{L}_r(\cdot)=\mathcal{S}(\cdot)$, rather than directly calculating the compressed model size. By explicitly building the relationship between compressed model size with lossless techniques and entropy term, we can guide the optimization to pursue a better weight distribution, which is more appropriate for later lossless compression. In this way, the potential of the subsequent lossless compression can be well unleashed.
\begin{equation}
\begin{aligned}
& \min_{\hat{\vw}} \E \left[
    \|\mathcal{F} (\vx, \hat{\vw})) - \mathcal{F} (\vx, \vw)\|_{F}^{2}
     + \lambda \cdot S(\hat{\vw})\right].
\end{aligned}
\end{equation}

By substituting \autoref{eq3} into the above equation, the following one can be deduced:
\begin{equation}\label{eqrd}
\min_{\hat{\vw}} \E \left[
    \|\mathcal{F} (\vx, \hat{\vw})) - \mathcal{F} (\vx, \vw)\|_{F}^{2}
    - \lambda \cdot \sum_{\hat{w} \in \hat{\vw}}log_2\mathcal{P}(\hat{w})\right].\\
\end{equation}

\vspace{1em}
\noindent\textbf{Compression ratio control.}
Furthermore, considering our objective is to achieve high accuracy while meeting the compression ratio requirement, we introduce $CR_{\text{target}}$ into the term to control the whole compression ratio as shown in \autoref{eq18}. Once the $CR_{\text{target}}$ is attained, the lossy compression will be subject only to its original objective:
\begin{equation}\label{eq18}
    \mathcal L_r(\hat{\vw})=ReLU(\sum_{\hat{w} \in \hat {\vw}}-log_2\mathcal{P}(\hat w)-\frac{32\cdot numel(\vw)}{CR_{target}}),
\end{equation}
where $32 \cdot numel({\vw})$ is the size of the original weights $\vw$.
\vspace{1em}

Optimizing the above equations is not straightforward. Specifically, to attain our optimization objective, we put forward a method that can jointly optimize various lossy compression methods with a unified weight transformation in \autoref{unified_weight_transformation}. Additionally, a method is presented to ensure the differentiability of the probability mass function by leveraging a kernel function in \autoref{subsec_differentiable_counter}.

\begin{figure}[t]
\begin{center}
\includegraphics[width=1\linewidth]{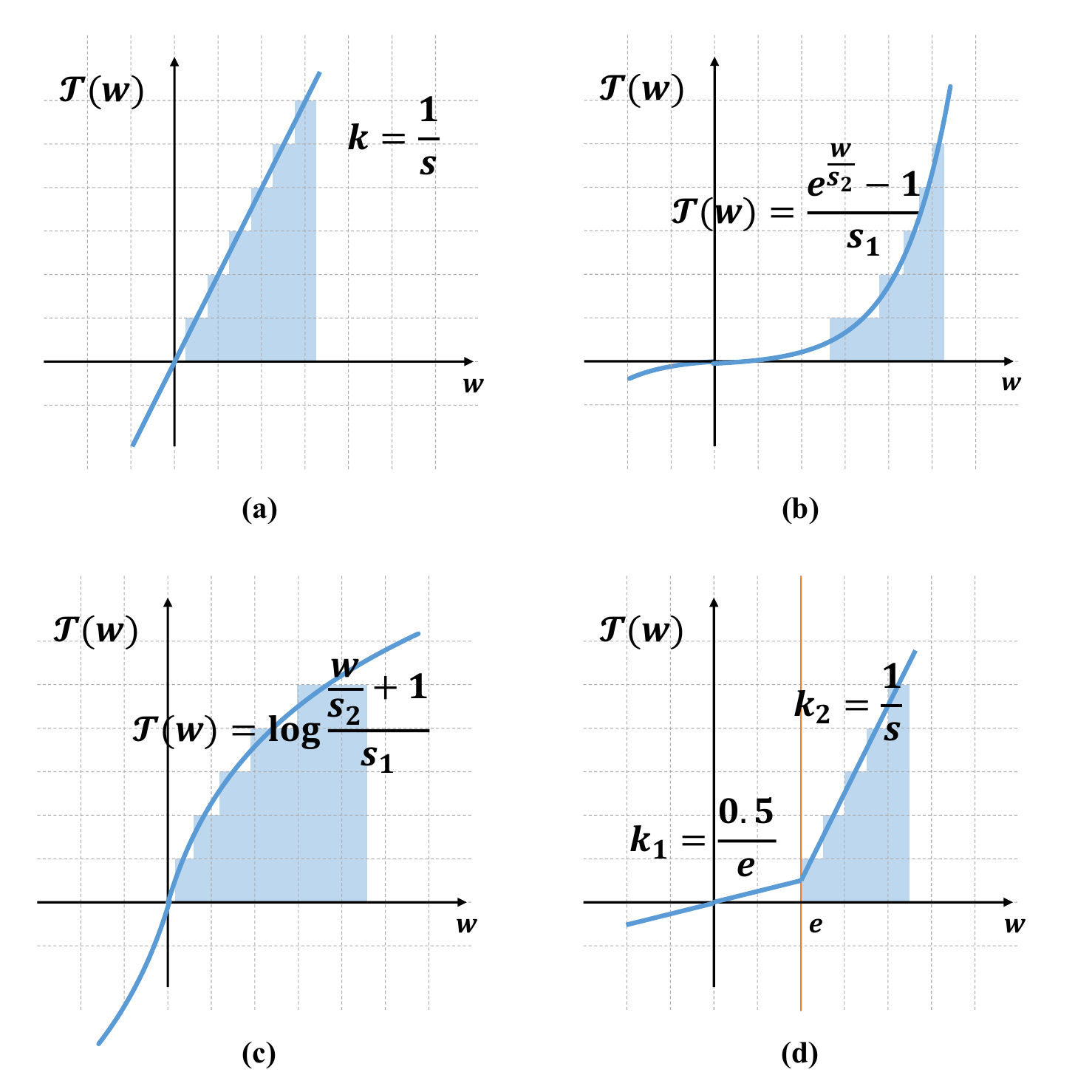}
\end{center}
   \caption{Some forms of transformations: (a) refers to linear quantization; (b) refers to exponential quantification; (c) refers to logarithmic quantization; (d) refers to joint pruning and quantization.}
\label{fig:function}
\end{figure}

\subsection{Unified Weight Transformation}
\label{unified_weight_transformation}

Previous studies \cite{DBLP:journals/corr/HanMD15, DBLP:conf/cvpr/TungM18} treated different compression methods separately without considering how they might affect each other. Or they lacked a unified approach to constrain compression methods according to the optimization objective. Consequently, they heavily relied on retraining the model's weights to achieve desired results, making it challenging to obtain satisfactory outcomes in a post-training setting. To address these issues, we abstract different lossy compression methods into a unified weight transformation.

\vspace{1em}
\noindent{\textbf{Unified representation.}}
Here, we propose to present quantization and pruning functions in a unified way. Quantization usually first quantizes weights into integers and then recovers them to floating-points, as defined in \autoref{eq_quantization}.
\begin{equation}
\label{eq_quantization}
\hat{w} =\mathcal{T}^{-1}(\lfloor\mathcal{T}(w)\rceil),
\end{equation}
where $\lfloor\cdot\rceil$ is the round-to-nearest operation, the element-wise function $\mathcal{T(\cdot)}$ encodes the weight transformation, and $\mathcal{T}^{-1}(\cdot)$ is its reverse function. Note that $\mathcal{T(\cdot)}$ always serves as a continuous function under different quantization settings, thus $\mathcal{T}^{-1}(\cdot)$ always exists. Especially, we can achieve uniform quantization by setting $\mathcal{T}(w) = \frac{w}{s}$, where $s$ stands for the interval of compressed data, as shown in \autoref{fig:function} (a). Non-uniform quantization like logarithmic or exponential ones can also be obtained by taking some non-linear $\mathcal{T(\cdot)}$ functions, as illustrated in \autoref{fig:function} (b) and (c).

Motivated by the quantization process \autoref{eq_quantization}, we propose to denote the pruning in the same way but with different $\mathcal{T(\cdot)}$ and $\mathcal{T}^{-1}(\cdot)$ functions. The $\mathcal{T(\cdot)}$ transformation for pruning can be devised as below:
\begin{equation}
    \mathcal{T}(w) =  \left\{
\begin{aligned}
&\frac{0.5}{e} \cdot w , &|w| < e \\
&w, &|w| \geq e
\end{aligned}
\right. ,
\label{eq10}
\end{equation}
where $e$ is the pruning threshold. In \autoref{eq10}, it can be observed that elements inside $(-e, e)$ are projected into $(-0.5, 0.5)$, which will become zero after rounding and reverse functions. Otherwise, values are kept intact. Therefore, unstructured pruning can be practiced by taking the same \autoref{eq_quantization} like quantization and assigning $\mathcal{T(\cdot)}$ to \autoref{eq10}. That is to say, unstructured pruning and different kinds of quantization can be clarified under the same concept.

\vspace{1em}
\noindent \textbf{Joint pruning and quantization.}
The unified representation \autoref{eq_quantization} for quantization and pruning brings us a way to integrate and optimize them together. By designing a continuous piece-wise function as shown in \autoref{equ11} and Figure \ref{fig:function} (d), we can pursue a joint objective.
\begin{equation}
\mathcal{T}(w) = \left\{
\begin{aligned}
&\frac{0.5}{e}\cdot w , &|w| < e \\
&sign(w)\cdot(\frac{|w|-e}{s}+0.5), &|w| \geq e
\end{aligned}
\right.,
\label{equ11}
\end{equation}
where weights whose absolute values are smaller than $e$ will be pruned, and rest non-pruned weights will be operated with $s$ to achieve quantization.

\vspace{1em}
\autoref{equ11} established a joint form for lossy compression. By using such a $\mathcal{T(\cdot)}$ and \autoref{eq_quantization} in \autoref{eq5}, joint pruning and quantization optimization can be realized. Note that apart from weight tuning for our objective, parameters of pruning and quantization methods (i.e, $e$ and $s$) are also optimized together. Consequently, by adopting $\mathcal{T}_i(\cdot)$ for each layer, we can explore various compression methods for each layer and determine their unique strategies. Thus, the model can be effectively compressed with enhanced outcomes in the post-training manner. 

\begin{table*}[t]
\small
\begin{center}
\begin{tabular*}{0.83\linewidth}{llcccc}
\toprule
Model                         &  Method                & Top1-acc & Top5-acc   & Target CR & CR \\ \midrule
\multirow{7}{*}{ResNet-18}    & uncompressed           & 71.06\% & 89.90\%     & -     & $1.0\times$      \\
                              & DFQ                    & 61.50\% & 83.56\%     & -     & $9.8\times$  \\
                              & DeepCABAC              & 70.68\% & 89.67\%     & -     & $7.6\times$  \\
                              & AdaRound               & 70.0\%   & 89.26\%      & -     & $9.7\times$       \\
                              & Data-Aware PNMQ* \ \   & \ \ 69.21\% / 69.76\%\ \  & \ \ 88.76\% / 89.08\%\ \  & - &$7.4\times$ \\          
                              & Ours                   & 70.79\% & 89.94\%     & $12\times$   & $12.0\times$ \\
                              & Ours                   & 70.20\% & 89.47\%     & $15\times$   & $15.0\times$ \\ \midrule
\multirow{6}{*}{ResNet-50}    & uncompressed           & 76.63\% & 93.07\%     & -     & $1.0\times$      \\
                              & DFQ                    & 75.47\% & 92.36\%     & -     & $6.1\times$  \\
                              & AdaRound               & 75.87\% & 92.66\%     & -     & $9.4\times$       \\   
                              & Data-Aware PNMQ*       & 75.5\% / 76.13\%  & 92.74\% / 92.86\% & - & $7.8\times$ \\
                              & Ours                   & 76.68\% & 93.13\%     & $10\times$   & $10.0\times$ \\
                              & Ours                   & 75.95\% & 92.82\%     & $15\times$   & $15.0\times$ \\ \midrule
\multirow{7}{*}{MobileNetV2}  & uncompressed           & 72.62\% & 90.62\%     & -    & $1.0\times$   \\
                              & DFQ                    & 68.70\% & 88.30\%     & -    & $5.0\times$   \\
                              & DeepCABAC              & 72.00\% & 90.39\%     & -    & $4.8\times$   \\
                              & AdaRound               & 69.01\% & 88.86\%     & -    & $7.2\times$   \\     
                              & Data-Aware PNMQ*       & 71.68\% / 71.88\% & 90.2\% / 90.29\% & - & $4.9\times$ \\
                              & Ours                   & 72.29\% & 90.58\%     & $8\times$   & $7.9\times$   \\
                              & Ours                   & 71.53\% & 90.18\%     & $10\times$  & $9.9\times$  \\ \midrule
\multirow{5}{*}{RegNet-600m}  & uncompressed           & 73.55\% & 91.57\%     & -    & $1.0\times$  \\
                              & DFQ                    & 73.32\% & 91.52\%     & -    & $5.6\times$  \\
                              & DeepCABAC              & 70.76\% & 90.49\%     & -    & $5.3\times$  \\
                              & AdaRound               & 71.92\% & 90.65\%     & -    & $9.7\times$  \\  
                              & Ours                   & 73.08\% & 91.25\%     & $12\times$  & $11.9\times$ \\ \midrule
\multirow{6}{*}{RegNet-3200m \ \ } & uncompressed           & 78.36\% & 94.16\%     & -    & $1.0\times$  \\
                              & DFQ                    & 78.06\% & 94.02\%     & -    & $5.8\times$  \\
                              & DeepCABAC              & 77.02\% & 93.40\%     & -    & $5.5\times$  \\
                              & AdaRound               & 77.32\% & 93.57\%     &      & $10.4\times$ \\   
                              & Ours                   & 78.32\% & 94.13\%     & $10\times$  & $10.0\times$ \\
                              & Ours                   & 77.82\% & 93.89\%     & $15\times$  & $15.0\times$ \\ \midrule
\multirow{5}{*}{MNasNet}      & uncompressed           & 76.56\% & 93.15\%     & -    & $1.0\times$   \\
                              & DFQ                    & 76.20\% & 92.97\%     & -    & $5.1\times$   \\
                              & DeepCABAC              & 74.04\% & 92.02\%     & -    & $5.1\times$   \\
                              & AdaRound               & 74.41\% & 91.99\%     &      & $8.4\times$   \\
                              & Ours                   & 76.04\% & 92.77\%     & $12\times$  & $12.0\times$  \\ \bottomrule                            
\end{tabular*}
\end{center}
\caption{
     Comparison results with state-of-the-arts on various networks. Our unified compression method achieves the best performance. (* means using the results published in the original paper.)
}
\label{tb1}
\end{table*}

\subsection{Differentiable Counter}
\label{subsec_differentiable_counter}
\begin{figure}[t]
\begin{center}
\includegraphics[width=1\linewidth]{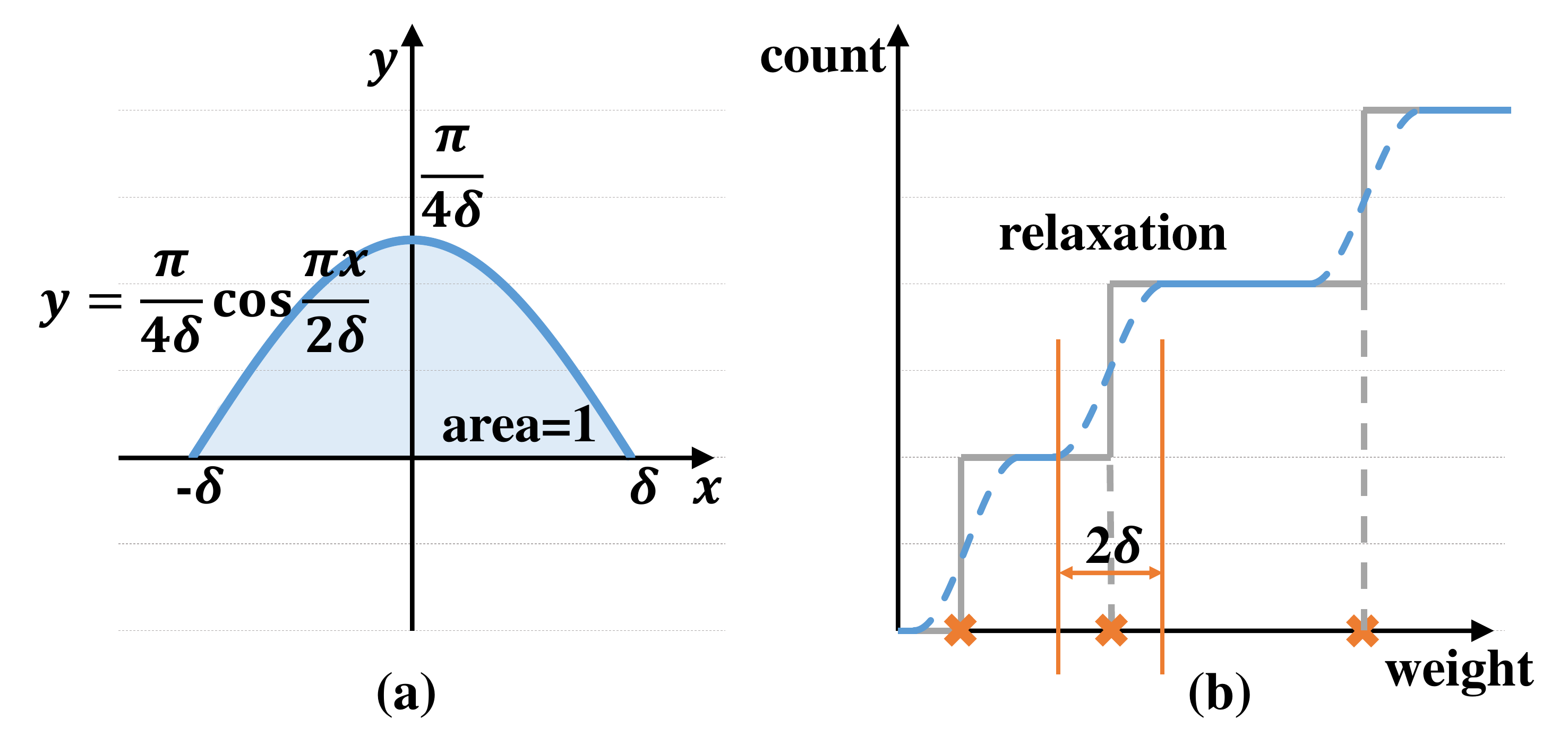}
\end{center}
   \caption{(a) refers to kernel function for relaxation; (b) refers to the relaxed count function.}
\label{fig:counter}
\end{figure}
To minimize \autoref{eqrd}, the differentiability of $\mathcal{P}(\hat{w})$ is required, as it needs to guide the learning of parameters in $\mathcal{T}(\cdot)$. Since $\hat{w}=\mathcal{T}^{-1}(\lfloor \bar{w} \rceil)$ and $\mathcal{T}^{-1}(\cdot)$ is bijective, we can deduce that $\mathcal{P}(\hat{w}) = \mathcal{P}(\lfloor \bar{w} \rceil)$. We model $\mathcal{P}(\lfloor \bar{w} \rceil)$ instead of $\mathcal{P}(\hat{w})$ as the interval remains fixed during the rounding operation.

In comparison with complex probability models, directly counting the frequencies of weights provides a more accurate representation of the weight distribution. Based on this, a novel differentiable counter is designed to estimate the probability mass function, facilitating the learning. We initially define the cumulative distribution function $f_{\bar{\vw}}(x)$ for weights $\bar{\vw}$ as follows:
\begin{equation}\label{eq12}
    f_{\bar{\vw}}(x) = \frac{\mathcal{C}(\bar{\vw},x)}{numel(\bar{\vw})},
\end{equation}
\begin{equation}\label{eq13}
    \mathcal{C}(\bar{\vw},x) = \sum_{\bar w\in \bar{\vw} }(\epsilon(x - \bar w)),
\end{equation}
\begin{equation}\label{eq14}
    \epsilon(x) = \left\{
        \begin{aligned}
        &0, &x < 0 \\
        &1, &x \geq 0 
        \end{aligned}
    \right.,
\end{equation}
where $numel(\bar{\vw})$ represents the number of weights in $\bar{\vw}$, and $\mathcal{C}(\bar{\vw},x)$ is a counter that counts the number of weights less than $x$ in $\bar{\vw}$. 

To ensure the differentiability of $\mathcal{C}(\bar{\vw},x)$, we leverage a kernel function and replace the step function $\epsilon(x)$. In detail, the relaxation function is defined as below:
\begin{equation}\label{eq15}
   \mathcal{C}(\bar{\vw},x) \cong \sum_{w\in \bar{\vw}}\int _{-\infty}^{x -w}\mathcal{K}(u) du,
\end{equation}
\begin{equation}\label{eq16}
    \mathcal{K}(x) = \left\{
        \begin{aligned}
        &0, &x \in (-\infty, \delta)\cup(\delta,\infty)\\
        &\frac{\pi}{4\delta} \cdot cos(\frac{\pi x}{2\delta}), &x \in [-\delta, \delta]
        \end{aligned}
    \right.,
\end{equation}
where $\mathcal{K}(\cdot)$ is the kernel function with a relaxation factor $\delta$ as shown in \autoref{fig:counter}. When $\delta$ approaches zero, the function simplifies to \autoref{eq13}. Considering the rounding operation, we estimate the probability mass of $\lfloor \bar{w} \rceil$ based on the difference in cumulative probability function between $\lfloor \bar w \rceil + 0.5$ and $\lfloor \bar w \rceil- 0.5$:
\begin{equation}\label{eq17}
\begin{aligned}
\mathcal{P}(\lfloor \bar w \rceil) 
= f_{\bar{\vw}}(\lfloor \bar w \rceil + 0.5) -  f_{\bar{\vw}}(\lfloor \bar w \rceil - 0.5).
\end{aligned}
\end{equation}

With the differentiable counter, obtaining the differentiable probability mass of the weights $\lfloor \bar{\vw} \rceil$ becomes easy. Besides, we utilize the Straight-Through Estimator (STE) \cite{DBLP:journals/corr/BengioLC13} to obtain gradients for rounding operations. Hence, the entropy term in \autoref{eqrd} can be minimized to achieve $\lfloor \bar{\vw} \rceil$ more suitable for later lossless compression during the weight transformation process, enabling a more effective combination of lossy and lossless compression.

\input{exp_final}

\section{Conclusion}
In this work, we propose a novel post-training compression method that combines lossy and lossless compression. For lossy compression, we unify the modeling of weight distortion via a unified weight transformation for pruning, quantization, and so on. Moreover, we design a dedicated differentiable counter that accurately computes the information entropy of the compressed weights, which can regulate the weights and adapt to later lossless compression, thus achieving a better compression ratio. Extensive experiments on various networks prove the superiority of our method compared to previous methods. Furthermore, our work provides a meaningful perspective for more extreme model compression in the future by unifying different compression methods.

\section*{Acknowledgment}
We sincerely thank the anonymous reviewers for their serious reviews and valuable suggestions to make this better. This work was supported in part by the National Natural Science Foundation of China under Grant 62206010 and 62072019.

{\small
\bibliographystyle{ieee_fullname}
\bibliography{egbib}
}

\appendix
\input{supplementary_final}

\end{document}

%% file: math_commands.tex
\usepackage{amsmath,amsfonts,bm}

\def\eqref#1{equation~\ref{#1}}
\def\1{\bm{1}}

\def\vw{{\bm{w}}}
\def\vx{{\bm{x}}}

\DeclareMathAlphabet{\mathsfit}{\encodingdefault}{\sfdefault}{m}{sl}
\SetMathAlphabet{\mathsfit}{bold}{\encodingdefault}{\sfdefault}{bx}{n}

\newcommand{\E}{\mathbb{E}}

%% file: exp_final.tex
\section{Experiment}
This section offers a comprehensive evaluation of our method. We compare and analyze the results of the method with state-of-the-art methods on the classification task, and present results on the detection task. In addition, ablation studies are conducted, accompanied by the analysis of time cost, weight transformation, and compression ratio.

\subsection{Experimental Setting}
\label{sec_exp_set}
To demonstrate the generality of our proposed method, extensive evaluations encompassing both the classification task and the object detection task are performed. For the classification task, we assess its performance across a wide range of convolutional neural networks, including ResNet \cite{DBLP:conf/cvpr/HeZRS16}, MobileNet \cite{DBLP:conf/cvpr/SandlerHZZC18}, RegNet \cite{DBLP:conf/cvpr/RadosavovicKGHD20}, and MNasNet \cite{DBLP:conf/cvpr/TanCPVSHL19}. The evaluation is carried out on the challenging ImageNet-1k dataset \cite{DBLP:conf/cvpr/DengDSLL009}, with a calibration set of 1000 images sampled from the training set to train the transformation parameters. Moreover, for the object detection task, we conduct experiments on the YOLOv5 model \cite{glenn_jocher_2021_5563715}, and the performance is validated on the widely used COCO2017 dataset \cite{DBLP:conf/eccv/LinMBHPRDZ14}.

We apply \autoref{equ11} for weight transformation and normal quantization for bias. Weight transformation parameters are initially trained, followed by fine-tuning of the weights. Inspired by knowledge distillation, both the complete network output and specific intermediate layer outputs contribute to the computation of the mean squared error loss. The networks are trained for 5 epochs, with each epoch containing 300 iterations of transformation training and 1000 iterations of weight fine-tuning. Finally, range coding, realized by \cite{bamler2022constriction}, is used to encode the model weights. More implementation details can be found in the supplementary material.

\subsection{Classification Task}
We compare our method with the state-of-the-art DFQ \cite{DBLP:conf/iccv/NagelBBW19} (only weight 8-bit), DeepCABAC \cite{DBLP:journals/corr/abs-1905-08318, DBLP:journals/jstsp/WiedemannKMHMMN20}, Data-Aware PNMQ \cite{DBLP:conf/cvpr/ChikinA22}, and AdaRoun \cite{DBLP:conf/icml/NagelABLB20} (only weight 4-bit) on ImageNet-1k. In the case of AdaRound, the first and last layers are 8-bit weights. Additionally, to ensure a fair comparison, we apply range coding for all methods.

\autoref{tb1} lists the performance of all methods. Thanks to the unified modeling of different compression techniques, our method could easily achieve the best compression ratio with little accuracy loss, proving the superiority of our method. For instance, the proposed method reduces the model size by more than 12 times on ResNet-18 with only a 0.3\% accuracy drop, achieving about 60\% improvement than PNMQ. Moreover, we can find that different networks have different sensitivities to compression. MobileNet-v2 is the most sensitive, and RegNet-3200m with a larger number of parameters has a higher compression potential.

\subsection{Object Detection Task}
\begin{figure}[ht]
\begin{center}
\includegraphics[width=1\linewidth]{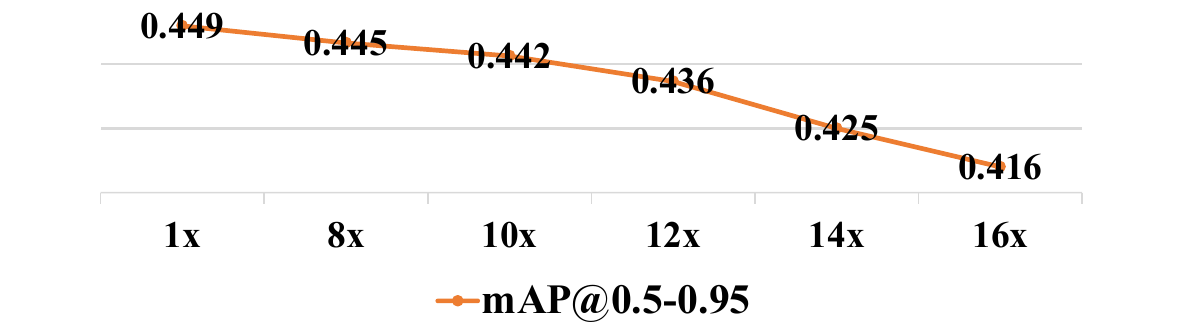}
\end{center}
   \caption{The mAP@0.5-0.95 of YOLOv5 with CR increasing.}
\label{fig_yolo}
\end{figure}
Besides, we conduct experiments on the obeject detection task, using a YOLOv5 model and the COCO2017 dataset. The final results are shown in \autoref{fig_yolo}. In \autoref{fig_yolo}, mAP (mean average precision) is a widely-adopted metric, which serves to evaluate the accuracy and recall rate of a model in detecting objects across different classes. A higher mAP implies that the model can better identify objects. It can be observed that our method still performs remarkably well on this task.

\begin{figure*}[h]
\begin{center}
\includegraphics[width=1\linewidth]{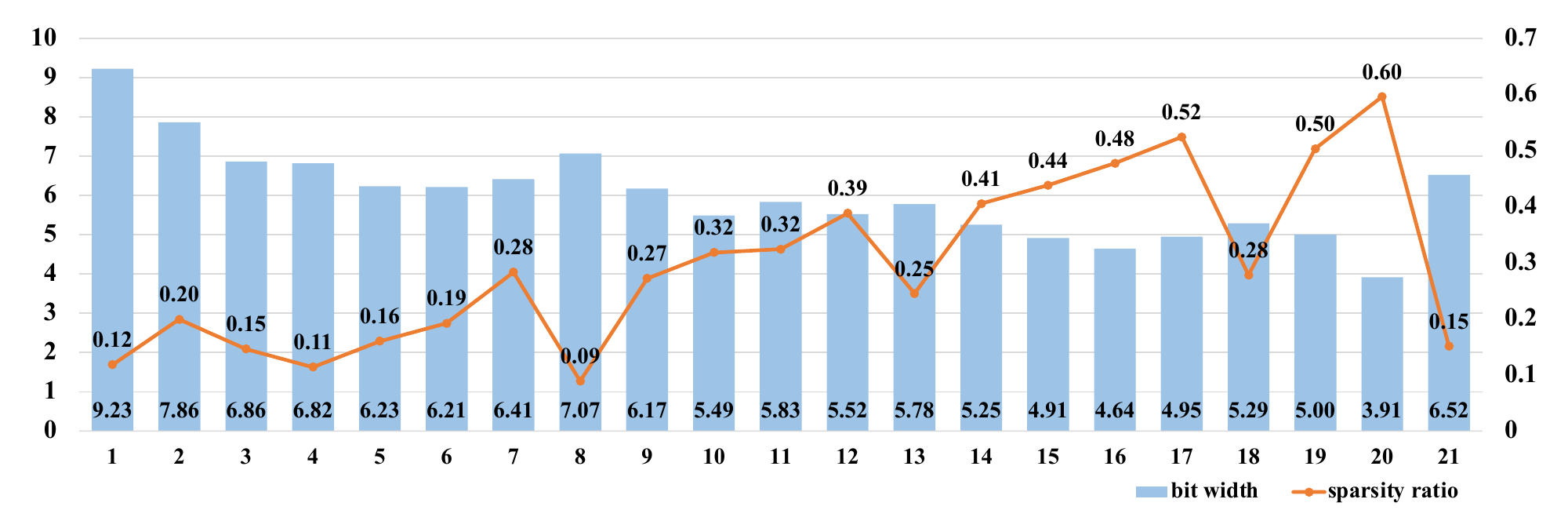}
\end{center}
   \caption{The distribution of bit width and sparsity ratio of ResNet-18 at a compression ratio of $20\times$.}
\label{fig_exp}
\end{figure*}
\subsection{Ablation Study}

In this part, ablation experiments will be performed on the proposed unified weight transformation $\mathcal{T}(\cdot)$ and the differentiable counter.

\vspace{1em}
\noindent\textbf{Choice of weight transformation.} 
\autoref{tab_trans} presents the results of various transformations described in Figure \autoref{fig:function}. It is evident that different transformations can achieve comparable performance when the compression ratio is close to $15\times$ on ResNet-18, demonstrating the robustness of our transformation. However, as the compression ratio increases to $20\times$, the transformation that combines pruning and quantization exhibits better performance. This observation highlights that our unified transformation effectively harnesses diverse compression methods, enabling the possibility of achieving extreme model compression.
\begin{table}[h]
\begin{center}
\begin{tabular}{lcc}
\toprule
weight transformation $\mathcal{T(\cdot)}$  & CR                   & Top1-acc          \\ \midrule
linear quantization               & $14.97\times$ & 70.37\%  \\ 
log quantization                  & $14.91\times$ & 70.24\%  \\ 
exp quantization                  & $14.94\times$ & 70.29\%  \\ 
joint pruning and quantization    & $14.94\times$ & 70.20\%  \\ \midrule
linear quantization               & $19.85\times$ & 68.34\%  \\ 
log quantization                  & $19.89\times$ & 68.47\%  \\ 
exp quantization                  & $19.87\times$ & 68.75\%  \\ 
joint pruning and quantization    & $19.95\times$ & 69.15\%  \\ \bottomrule
\end{tabular}
\end{center}
\caption{\centering Compression results with different transformations.}
\label{tab_trans}
\end{table}

% \vspace{1em}
\noindent\textbf{Choice of differentiable counter.} \autoref{tab_abl} provides compression results on ResNet-18 at $15\times$ CR with different relaxation factors $\delta=\frac{max(\vw) - min(\vw)}{resolution}$ and kernel functions. We explored three kernel functions (cosine, linear, and triangle) and four resolutions (16, 32, 64, 128) for the differentiable counter. The results under all settings demonstrate the robustness of our differentiable counter.
% \vspace{1em}

\begin{table}[t]
\begin{center}
\begin{tabular}{lcccc}
\toprule
kernel               & resolution & CR     & Top1-acc   & Top5-acc   \\ \midrule
\multirow{4}{*}{cosine} & 16         & $15.10\times$ & 70.14\% & 89.62\% \\ 
                     & 32         & $15.03\times$ & 70.25\% & 89.60\% \\ 
                     & 64         & $14.93\times$ & 70.21\% & 89.60\% \\  
                     & 128        & $15.00\times$ & 70.04\% & 89.63\% \\ \midrule
\multirow{4}{*}{linear} & 16         & $15.29\times$ & 70.15\% & 89.57\% \\ 
                     & 32         & $15.15\times$ & 70.05\% & 89.40\% \\ 
                     & 64         & $14.95\times$ & 70.07\% & 89.50\% \\
                     & 128        & $15.00\times$ & 70.20\% & 89.57\% \\ \midrule
\multirow{4}{*}{triangle} & 16         & $15.06\times$ & 70.11\% & 89.56\% \\ 
                     & 32         & $15.11\times$ & 70.09\% & 89.52\% \\ 
                     & 64         & $15.09\times$ & 70.14\% & 89.61\% \\ 
                     & 128        & $14.90\times$ & 70.14\% & 89.64\% \\ \bottomrule
\end{tabular}
\end{center}
\caption{
    Performance on ResNet-18 at $15\times$ CR for the differentiable counter with different resolutions and kernel functions.
}
\label{tab_abl}
\end{table}

\begin{figure}[t]
\begin{center}
\includegraphics[width=1\linewidth]{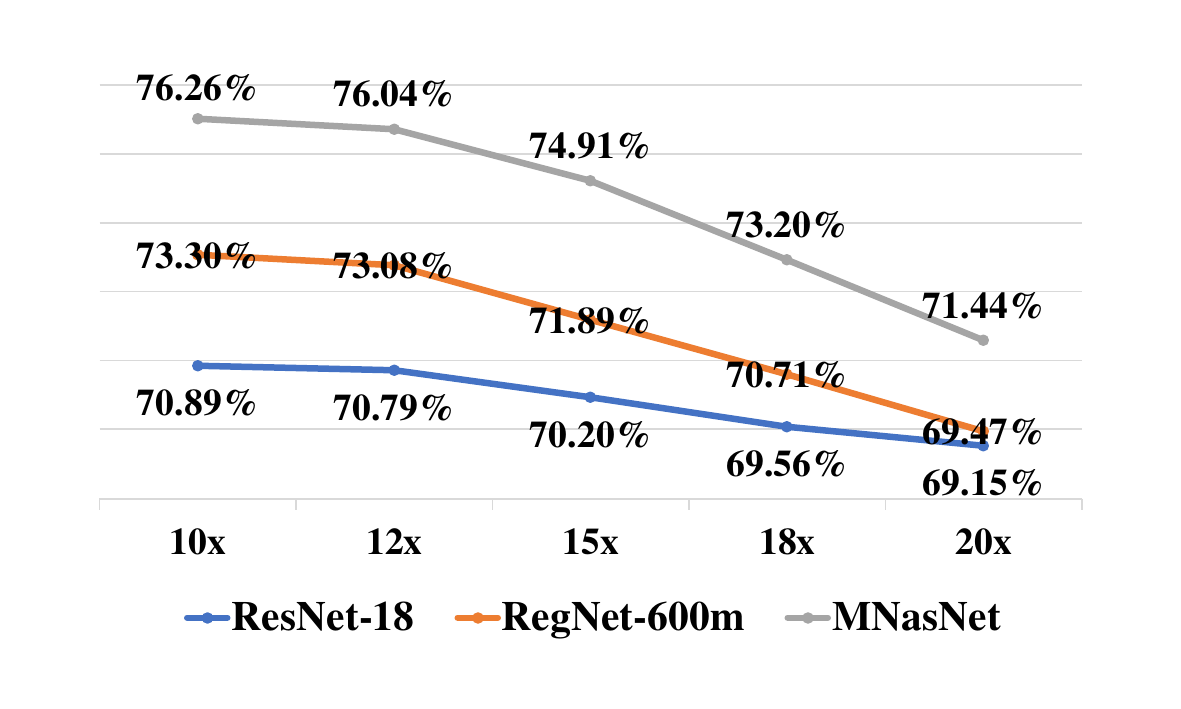}
\end{center}
   \caption{The accuracy of compressed models with CR increasing.}
\label{fig_CR}
\end{figure}

\subsection{Analysis}
In this part, we provide some analysis of our method.

\vspace{1em}
\noindent\textbf{Analysis of time consumption.}
The main time cost of our method during compression comes from the training of transformation parameters and weight fine-tuning. We use a V100 GPU with 16GB memory and set the batch size to 32. For models such as ResNet-18, the training takes around 0.8s per iteration, while fine-tuning takes around 0.08s. Larger models may take more time. Typically, 500 iterations for training and 2000 for fine-tuning can achieve satisfactory results, which take only around 10 minutes.

As for the overhead of decompression during inference, lossless compression is applied to the weights after the complete transformation $\mathcal{T}^{-1}(\lfloor\mathcal{T}(w)\rceil)$. Hence, we only need to decode the entropy coding without dequantization or any other inverse transformations. The time cost of decoding is worth considering, but it can be mitigated by employing various techniques, such as heterogeneous computing.

\vspace{1em}
\noindent\textbf{Analysis of weight transformation.}
The transformation combining quantization and pruning adaptively selects suitable quantization bits and sparsity ratios for each layer, as \autoref{fig_exp} shows. It helps detect layers with redundant parameters, such as ``layer4.1.conv2" (20 in the figure) in ResNet-18, meaningful for network structure optimizing.

\vspace{1em}
\noindent\textbf{Analysis of compression ratio.} 
\autoref{fig_CR} shows the trend of performance with respect to the compression ratio. It can be observed that, upon reaching a certain compression ratio, such as $12\times$ on MNasNet, the model experiences a significant accuracy drop. The controllability of the compression ratio in our method assists us in a more practical trade-off between accuracy and compression ratio.

%% file: supplementary_final.tex
\section{Implementation Details}
This section primarily introduces the selection of distillation layers and the implementation of the differentiable counter.

\subsection{Fine-tuning Knowledge Distillation Layers}
As mentioned in the previous \autoref{sec_exp_set}, we select certain outputs of the intermediate layers for distillation during the training and fine-tuning process. Specifically, we choose the output of each block of the models as knowledge distillation layers, taking into account both the stability of the model's convergence and its final compression performance during our experiments.

\subsection{Differentiable Counter}
In the process of implementing the differentiable counter, the weights $\hat \vw$ are first sorted, and then binary search is used to determine the position of the query value $x$ in the sorted weights. By using relaxation, the forward propagation value can be obtained. In the process of calculating gradients during backward propagation, it is important to take note of the accumulation of associated gradients. In practice, weight sampling is used to count the frequencies of values when the number of parameters in a layer exceeds a certain amount (such as 2 to the power of 14) due to the relatively slow traversal of counts. We have observed that the utilization of sampling leads to a substantial improvement in computational efficiency while maintaining a satisfactory level of accuracy. To this end, we offer two implementations: one in C++ and the other in CUDA \cite{cuda}.

\section{Additional Experimental Results}
In this section, we present additional experimental results and analyses that were not included in the main text. These results include the analysis of compression ratio allocations for layers in more networks, ablation experiments on encoding algorithms, calibration dataset size, and weight-finetuning, as well as experiments on a novel network architecture.

\subsection{Compression Ratio Allocation for Layers}
In this section, we will present more distributions of bit width and sparsity ratio for the compressed models, which were not shown in the experiment section. We have selected one representative model each from MobileNet, RegNet, and MNasNet. Since these models exhibit varying levels of accuracy at different compression ratios, we have chosen the highest compression ratios possible while maintaining an acceptable level of accuracy.

\begin{figure*}[!htb]
\begin{center}
\includegraphics[width=1.0\linewidth]{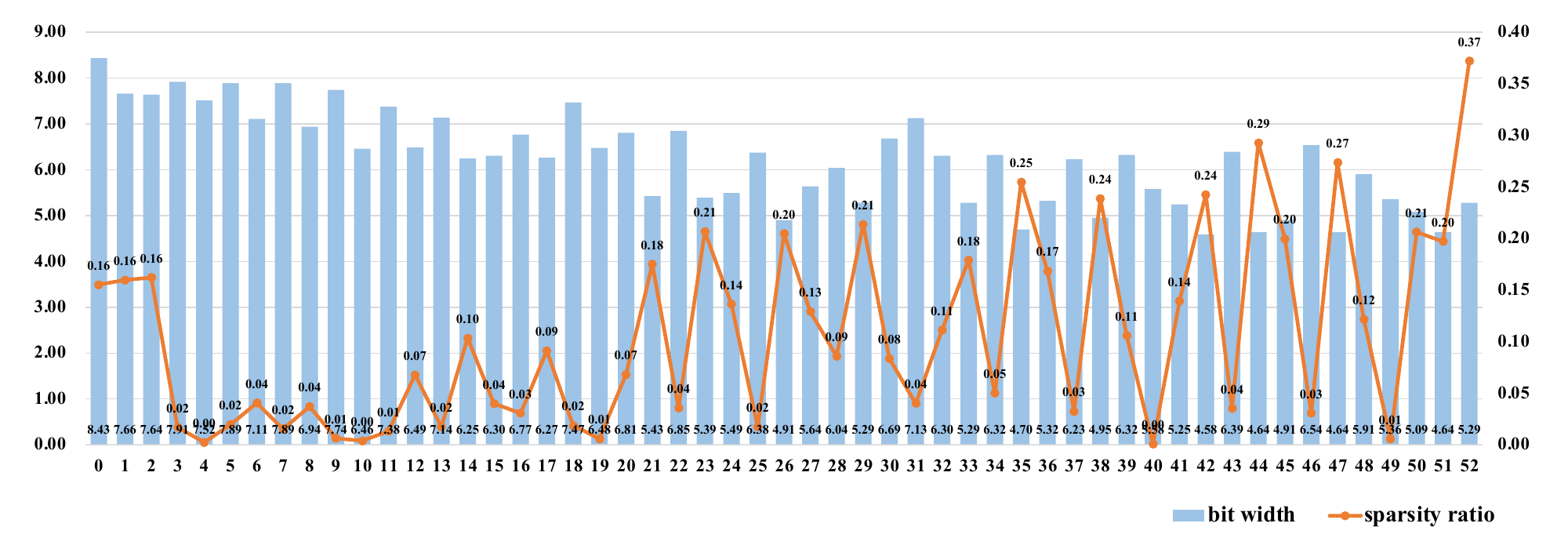}
\end{center}
   \caption{The distribution of bit width and sparsity ratio of MobileNetV2 at a compression ratio of $10\times$.}
\label{fig_exp1}
\end{figure*}

\vspace{1em}
\noindent {\bf MobileNetV2.}
\autoref{fig_exp1} depicts the distribution of bit width and sparsity ratio for MobileNetV2 at a $10\times$ compression ratio. Our observations indicate that the earlier layers of the network generally have higher bit width values, while the later layers have lower bit width values. Additionally, sparsity is rarely observed in the earlier layers, but alternates between high and low ratios in the later layers. Remarkably, the final layer, which is a classifier, exhibits the highest sparsity ratio across the entire network, which contrasts sharply with the characteristics of other models.

\begin{figure*}[!htb]
\begin{center}
\includegraphics[width=1.0\linewidth]{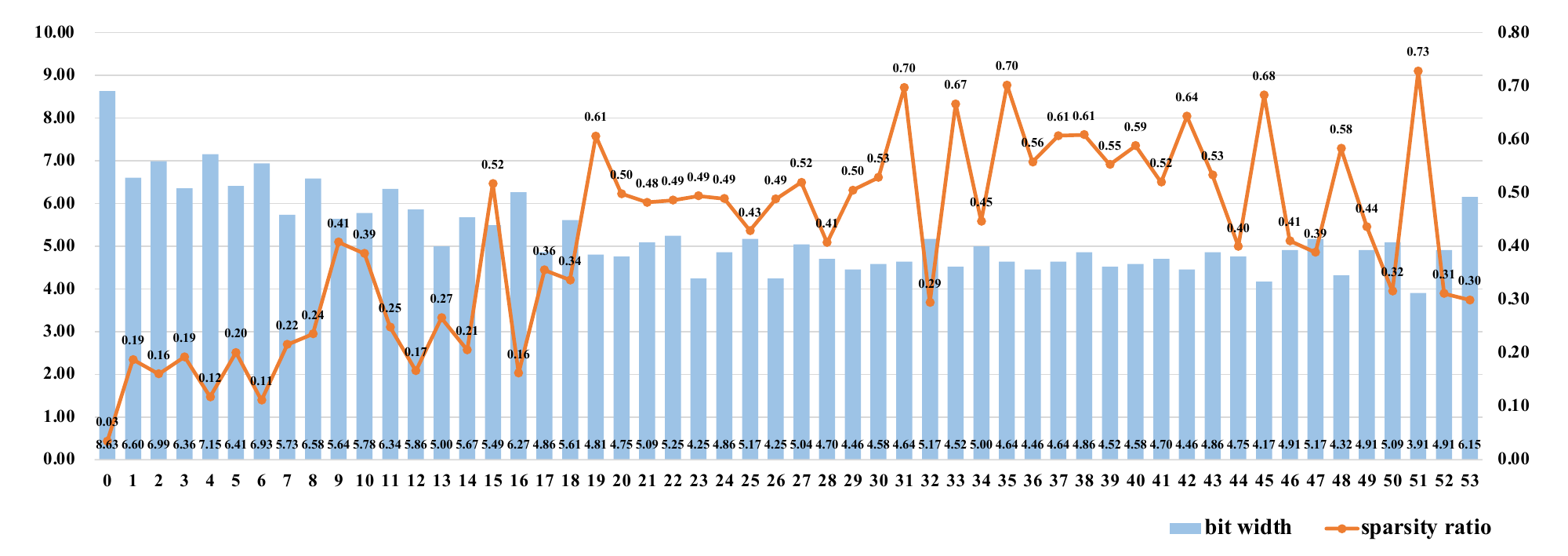}
\end{center}
   \caption{The distribution of bit width and sparsity ratio of RegNet-600m at a compression ratio of $15\times$.}
\label{fig_exp2}
\end{figure*}

\vspace{1em}
\noindent {\bf RegNet-600m.}
\autoref{fig_exp2} displays the distribution of bit width and sparsity ratio for RegNet-600m at a $15\times$ compression ratio. Our observations show that the compression ratio is more pronounced in the later layers of RegNet-600m, with lower bit width values and higher sparsity ratios. Conversely, the earlier layers exhibit lower compression ratios with lower sparsity ratios and higher bit width values. Meanwhile, the compression ratio of the classifier layer remains relatively low.

\begin{figure*}[ht]
\begin{center}
\includegraphics[width=1.0\linewidth]{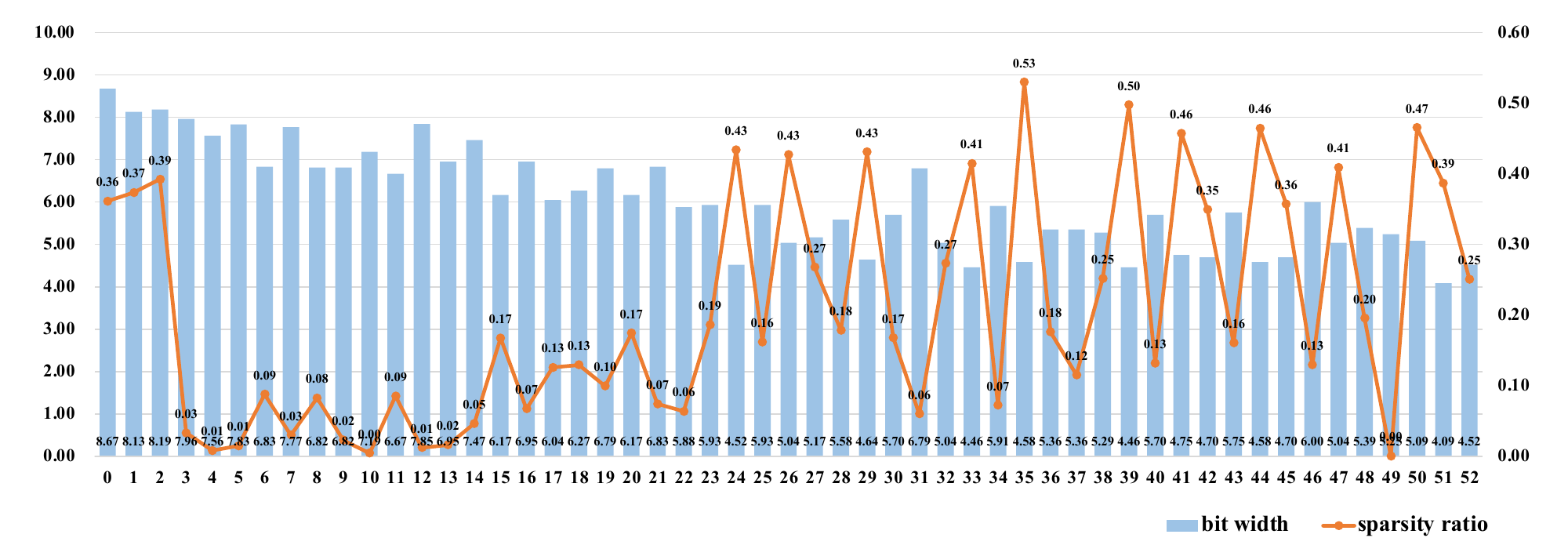}
\end{center}
   \caption{The distribution of bit width and sparsity ratio of MNasNet at a compression ratio of $12\times$.}
\label{fig_exp3}
\end{figure*}

\vspace{1em}
\noindent {\bf MNasNet.}
\autoref{fig_exp3} presents the result of MNasNet at a compression ratio of $12\times$. It is noteworthy that the layers with lower bit width values are located towards the end of the network, whereas the layers with higher bit width values are situated towards the front. This discrepancy is more pronounced compared to other networks.

\subsection{Additional Ablation Study}
In this section, we will provide some additional ablation studies, including the choice of lossless compression methods, the impact of calibrating dataset length, and the effect of weight fine-tuning.

\begin{table}[t]
\begin{center}
\begin{tabular}{lcc}
\toprule
Entropy Loss           & Huffman Coding        & Range Coding   \\ \midrule
4517.31KB      & 4588.33KB & 4527.30KB  \\ 
3719.25KB      & 3799.03KB & 3721.24KB  \\
3033.46KB      & 3130.71KB & 3040.09KB  \\
2496.27KB      & 2669.75KB & 2508.25KB  \\
2277.14KB      & 2500.24KB & 2288.99KB \\
\bottomrule
\end{tabular}
\end{center}
\caption{Effect of different coding methods on ResNet-18.}
\label{table_coding}
\end{table}
\vspace{1em}
\noindent {\bf Choice of lossless compression.}
We demonstrate the impact of using different encoding algorithms on the compression results, mainly comparing Huffman coding with range coding. \autoref{table_coding} presents the theoretical compression values calculated using the entropy regularization term under different target compression ratios for ResNet-18, the compressed model sizes after using Huffman coding, and the compressed model size after using range coding. It is apparent that the results of range coding are closer to the theoretical compression values, especially when the compression ratios are high. In such scenarios, the difference between the two coding methods is significant, and the advantages of range coding are more pronounced.

The reason lies in the fact that Huffman coding is a symbol coding method that uses only integer bits for encoding, which imposes certain limitations on the compressed data. In contrast, range coding does not suffer from such limitations and, as a result, offers more potential for achieving higher compression ratios. This is because range coding uses fractional bits for encoding, allowing for a more precise representation of the data and reducing the number of bits needed to represent the same information.

\vspace{1em}
\noindent {\bf Choice of calibration dataset size.}
In \autoref{tab_cal_size}, we observe that in the ImageNet-1k classification task, when the number of images in the calibration dataset is less than 768, the model's accuracy drop is more severe. However, when the number of images is greater than 768, the model's performance does not show significant improvement. Nevertheless, using more data for calibration would have negative impacts on calibration time and data acquisition. Therefore, we have opted to use 1000 images as the calibration dataset, striking a balance between different aspects.

\begin{table}[t]
\begin{center}
\begin{tabular}{lccccc}
\toprule
Length       &2000     &1000     &768     &512    &256      \\ \midrule
Accuracy/\%  &70.78    &70.79    &70.75   &69.79  &67.93    \\
\bottomrule
\end{tabular}
\end{center}
\caption{The results for ResNet-18 at $12\times$ compression ratio with different calibration sizes.}
\label{tab_cal_size}
\end{table}

\vspace{1em}
\noindent {\bf Effect of weight fine-tuning.}
We evaluate the performance of our unified compression method without weight fine-tuning, as shown in \autoref{table_finetune}. Our results demonstrate that our method can maintain a compression ratio of $10\times$ with minimal accuracy drop, solely relying on the unified transformation $\mathcal{T}(\cdot)$. This approach is still superior to existing state-of-the-art methods, highlighting the rationality and robustness of our proposed method.

\begin{table}[]
\begin{center}
\begin{tabular}{lcc}
\toprule
Model                                  & CR     & Top1-acc       \\ \midrule
\multirow{1}{*}{ResNet-18}             & $10.05\times$ & 70.47\%  \\  
\multirow{1}{*}{ResNet-50}             & $10.04\times$ & 76.22\%  \\  
\multirow{1}{*}{MobileNetV2}           & $9.864\times$ & 71.46\%  \\  
\multirow{1}{*}{RegNet-3200m}\ \ \ \ \ & \ \ $9.935\times$ \ \ & 78.00\%  \\ 
\multirow{1}{*}{RegNet-600m}           & $9.903\times$ & 72.82\%  \\ 
\multirow{1}{*}{MNasNet}               & $9.980\times$ & 75.78\%  \\ \bottomrule
\end{tabular}
\end{center}
\caption{The results without weight fine-tuning.}
\label{table_finetune}
\end{table}

\subsection{Performance on Vision Transformer}
\begin{table}[ht]
\begin{center}
\begin{tabular}{lcc}
\toprule
Target CR      & CR            & Top-1 acc   \\ \midrule
-              & $1.00\times$         & 85.09\%     \\
$6\times$             & $6.67\times$         & 84.97\%     \\  
$8\times$             & $8.05\times$         & 84.83\%     \\
$10\times$            & \ \ $10.05\times$ \ \ & 84.36\%     \\
$12\times$            & $12.01\times$        & 83.44\%     \\
$14\times$            & $14.02\times$        & 81.86\%     \\
\bottomrule
\end{tabular}
\end{center}
\caption{ViT compression results.}
\label{table_vit}
\end{table}

We also validated our method on the popular Vision Transformer network \cite{DBLP:conf/iclr/DosovitskiyB0WZ21}. Specifically, we conduct classification experiments on the ViT-base-patch16 using the same settings as the previous experiments. However, considering the computational complexity of the model and the need for efficiency, we set the batch size to 16 and train for 3 epochs, with 300 iterations for training transformation parameters per epoch and 1000 iterations for fine-tuning weights. As shown in \autoref{table_vit}, our method still achieves good performance on this network.

%% file: egpaper_final.bbl
\begin{thebibliography}{10}\itemsep=-1pt

\bibitem{bamler2022constriction}
Robert Bamler.
\newblock Understanding entropy coding with asymmetric numeral systems (ans): a
  statistician's perspective.
\newblock {\em arXiv preprint arXiv:2201.01741}, 2022.

\bibitem{DBLP:journals/jmlr/BaskinCZBBM21}
Chaim Baskin, Brian Chmiel, Evgenii Zheltonozhskii, Ron Banner, Alex~M.
  Bronstein, and Avi Mendelson.
\newblock {CAT:} compression-aware training for bandwidth reduction.
\newblock {\em J. Mach. Learn. Res.}, 22:269:1--269:20, 2021.

\bibitem{DBLP:journals/corr/BengioLC13}
Yoshua Bengio, Nicholas L{\'{e}}onard, and Aaron~C. Courville.
\newblock Estimating or propagating gradients through stochastic neurons for
  conditional computation.
\newblock {\em CoRR}, abs/1308.3432, 2013.

\bibitem{DBLP:conf/cvpr/ChikinA22}
Vladimir Chikin and Mikhail Antiukh.
\newblock Data-free network compression via parametric non-uniform mixed
  precision quantization.
\newblock In {\em {IEEE/CVF} Conference on Computer Vision and Pattern
  Recognition, {CVPR} 2022, New Orleans, LA, USA, June 18-24, 2022}, pages
  450--459. {IEEE}, 2022.

\bibitem{DBLP:conf/ijcnn/ChmielBZBYKBM20}
Brian Chmiel, Chaim Baskin, Evgenii Zheltonozhskii, Ron Banner, Yevgeny
  Yermolin, Alex Karbachevsky, Alex~M. Bronstein, and Avi Mendelson.
\newblock Feature map transform coding for energy-efficient {CNN} inference.
\newblock In {\em 2020 International Joint Conference on Neural Networks,
  {IJCNN} 2020, Glasgow, United Kingdom, July 19-24, 2020}, pages 1--9. {IEEE},
  2020.

\bibitem{DBLP:conf/iclr/ChoiEL17}
Yoojin Choi, Mostafa El{-}Khamy, and Jungwon Lee.
\newblock Towards the limit of network quantization.
\newblock In {\em 5th International Conference on Learning Representations,
  {ICLR} 2017, Toulon, France, April 24-26, 2017, Conference Track
  Proceedings}. OpenReview.net, 2017.

\bibitem{DBLP:conf/cvpr/DengDSLL009}
Jia Deng, Wei Dong, Richard Socher, Li{-}Jia Li, Kai Li, and Li Fei{-}Fei.
\newblock Imagenet: {A} large-scale hierarchical image database.
\newblock In {\em 2009 {IEEE} Computer Society Conference on Computer Vision
  and Pattern Recognition {(CVPR} 2009), 20-25 June 2009, Miami, Florida,
  {USA}}, pages 248--255. {IEEE} Computer Society, 2009.

\bibitem{DBLP:conf/nips/DongYAGMK20}
Zhen Dong, Zhewei Yao, Daiyaan Arfeen, Amir Gholami, Michael~W. Mahoney, and
  Kurt Keutzer.
\newblock {HAWQ-V2:} hessian aware trace-weighted quantization of neural
  networks.
\newblock In Hugo Larochelle, Marc'Aurelio Ranzato, Raia Hadsell,
  Maria{-}Florina Balcan, and Hsuan{-}Tien Lin, editors, {\em Advances in
  Neural Information Processing Systems 33: Annual Conference on Neural
  Information Processing Systems 2020, NeurIPS 2020, December 6-12, 2020,
  virtual}, 2020.

\bibitem{DBLP:conf/iccv/DongYGMK19}
Zhen Dong, Zhewei Yao, Amir Gholami, Michael~W. Mahoney, and Kurt Keutzer.
\newblock {HAWQ:} hessian aware quantization of neural networks with
  mixed-precision.
\newblock In {\em 2019 {IEEE/CVF} International Conference on Computer Vision,
  {ICCV} 2019, Seoul, Korea (South), October 27 - November 2, 2019}, pages
  293--302. {IEEE}, 2019.

\bibitem{DBLP:conf/iclr/DosovitskiyB0WZ21}
Alexey Dosovitskiy, Lucas Beyer, Alexander Kolesnikov, Dirk Weissenborn,
  Xiaohua Zhai, Thomas Unterthiner, Mostafa Dehghani, Matthias Minderer, Georg
  Heigold, Sylvain Gelly, Jakob Uszkoreit, and Neil Houlsby.
\newblock An image is worth 16x16 words: Transformers for image recognition at
  scale.
\newblock In {\em 9th International Conference on Learning Representations,
  {ICLR} 2021, Virtual Event, Austria, May 3-7, 2021}. OpenReview.net, 2021.

\bibitem{glenn_jocher_2021_5563715}
Glenn~Jocher et. al.
\newblock {ultralytics/yolov5: v6.0 - YOLOv5n 'Nano' models, Roboflow
  integration, TensorFlow export, OpenCV DNN support}, Oct. 2021.

\bibitem{DBLP:journals/corr/HanMD15}
Song Han, Huizi Mao, and William~J. Dally.
\newblock Deep compression: Compressing deep neural network with pruning,
  trained quantization and huffman coding.
\newblock In Yoshua Bengio and Yann LeCun, editors, {\em 4th International
  Conference on Learning Representations, {ICLR} 2016, San Juan, Puerto Rico,
  May 2-4, 2016, Conference Track Proceedings}, 2016.

\bibitem{DBLP:journals/corr/HanPTD15}
Song Han, Jeff Pool, John Tran, and William~J. Dally.
\newblock Learning both weights and connections for efficient neural networks.
\newblock {\em CoRR}, abs/1506.02626, 2015.

\bibitem{DBLP:conf/cvpr/HeZRS16}
Kaiming He, Xiangyu Zhang, Shaoqing Ren, and Jian Sun.
\newblock Deep residual learning for image recognition.
\newblock In {\em 2016 {IEEE} Conference on Computer Vision and Pattern
  Recognition, {CVPR} 2016, Las Vegas, NV, USA, June 27-30, 2016}, pages
  770--778. {IEEE} Computer Society, 2016.

\bibitem{DBLP:conf/eccv/HeLLWLH18}
Yihui He, Ji Lin, Zhijian Liu, Hanrui Wang, Li{-}Jia Li, and Song Han.
\newblock {AMC:} automl for model compression and acceleration on mobile
  devices.
\newblock In Vittorio Ferrari, Martial Hebert, Cristian Sminchisescu, and Yair
  Weiss, editors, {\em Computer Vision - {ECCV} 2018 - 15th European
  Conference, Munich, Germany, September 8-14, 2018, Proceedings, Part {VII}},
  volume 11211 of {\em Lecture Notes in Computer Science}, pages 815--832.
  Springer, 2018.

\bibitem{DBLP:journals/corr/HintonVD15}
Geoffrey~E. Hinton, Oriol Vinyals, and Jeffrey Dean.
\newblock Distilling the knowledge in a neural network.
\newblock {\em CoRR}, abs/1503.02531, 2015.

\bibitem{DBLP:conf/cvpr/HouQSMYXC00K22}
Zejiang Hou, Minghai Qin, Fei Sun, Xiaolong Ma, Kun Yuan, Yi Xu, Yen{-}Kuang
  Chen, Rong Jin, Yuan Xie, and Sun{-}Yuan Kung.
\newblock {CHEX:} channel exploration for {CNN} model compression.
\newblock In {\em {IEEE/CVF} Conference on Computer Vision and Pattern
  Recognition, {CVPR} 2022, New Orleans, LA, USA, June 18-24, 2022}, pages
  12277--12288. {IEEE}, 2022.

\bibitem{DBLP:journals/corr/abs-2102-07725}
Berivan Isik, Kristy Choi, Xin Zheng, Tsachy Weissman, Stefano Ermon,
  H.{-}S.~Philip Wong, and Armin Alaghi.
\newblock Neural network compression for noisy storage devices.
\newblock {\em CoRR}, abs/2102.07725, 2021.

\bibitem{DBLP:journals/corr/abs-2207-11048}
Andrey Kuzmin, Mart van Baalen, Markus Nagel, and Arash Behboodi.
\newblock Quantized sparse weight decomposition for neural network compression.
\newblock {\em CoRR}, abs/2207.11048, 2022.

\bibitem{DBLP:conf/iccvw/LazarevichKM21}
Ivan Lazarevich, Alexander Kozlov, and Nikita Malinin.
\newblock Post-training deep neural network pruning via layer-wise calibration.
\newblock In {\em {IEEE/CVF} International Conference on Computer Vision
  Workshops, {ICCVW} 2021, Montreal, BC, Canada, October 11-17, 2021}, pages
  798--805. {IEEE}, 2021.

\bibitem{DBLP:conf/nips/LiebenweinMFR21}
Lucas Liebenwein, Alaa Maalouf, Dan Feldman, and Daniela Rus.
\newblock Compressing neural networks: Towards determining the optimal
  layer-wise decomposition.
\newblock In Marc'Aurelio Ranzato, Alina Beygelzimer, Yann~N. Dauphin, Percy
  Liang, and Jennifer~Wortman Vaughan, editors, {\em Advances in Neural
  Information Processing Systems 34: Annual Conference on Neural Information
  Processing Systems 2021, NeurIPS 2021, December 6-14, 2021, virtual}, pages
  5328--5344, 2021.

\bibitem{DBLP:conf/eccv/LinMBHPRDZ14}
Tsung{-}Yi Lin, Michael Maire, Serge~J. Belongie, James Hays, Pietro Perona,
  Deva Ramanan, Piotr Doll{\'{a}}r, and C.~Lawrence Zitnick.
\newblock Microsoft {COCO:} common objects in context.
\newblock In David~J. Fleet, Tom{\'{a}}s Pajdla, Bernt Schiele, and Tinne
  Tuytelaars, editors, {\em Computer Vision - {ECCV} 2014 - 13th European
  Conference, Zurich, Switzerland, September 6-12, 2014, Proceedings, Part
  {V}}, volume 8693 of {\em Lecture Notes in Computer Science}, pages 740--755.
  Springer, 2014.

\bibitem{DBLP:journals/corr/abs-2109-07865}
Yuexiao Ma, Taisong Jin, Xiawu Zheng, Yan Wang, Huixia Li, Guannan Jiang, Wei
  Zhang, and Rongrong Ji.
\newblock {OMPQ:} orthogonal mixed precision quantization.
\newblock {\em CoRR}, abs/2109.07865, 2021.

\bibitem{DBLP:conf/icml/NagelABLB20}
Markus Nagel, Rana~Ali Amjad, Mart van Baalen, Christos Louizos, and Tijmen
  Blankevoort.
\newblock Up or down? adaptive rounding for post-training quantization.
\newblock In {\em Proceedings of the 37th International Conference on Machine
  Learning, {ICML} 2020, 13-18 July 2020, Virtual Event}, volume 119 of {\em
  Proceedings of Machine Learning Research}, pages 7197--7206. {PMLR}, 2020.

\bibitem{DBLP:conf/iccv/NagelBBW19}
Markus Nagel, Mart van Baalen, Tijmen Blankevoort, and Max Welling.
\newblock Data-free quantization through weight equalization and bias
  correction.
\newblock In {\em 2019 {IEEE/CVF} International Conference on Computer Vision,
  {ICCV} 2019, Seoul, Korea (South), October 27 - November 2, 2019}, pages
  1325--1334. {IEEE}, 2019.

\bibitem{cuda}
NVIDIA, Péter Vingelmann, and Frank~H.P. Fitzek.
\newblock Cuda, release: 10.2.89, 2020.

\bibitem{DBLP:conf/iclr/OktayBSS20}
Deniz Oktay, Johannes Ball{\'{e}}, Saurabh Singh, and Abhinav Shrivastava.
\newblock Scalable model compression by entropy penalized reparameterization.
\newblock In {\em 8th International Conference on Learning Representations,
  {ICLR} 2020, Addis Ababa, Ethiopia, April 26-30, 2020}. OpenReview.net, 2020.

\bibitem{DBLP:journals/corr/abs-2302-05397}
Nilesh~Prasad Pandey, Markus Nagel, Mart van Baalen, Yin Huang, Chirag Patel,
  and Tijmen Blankevoort.
\newblock A practical mixed precision algorithm for post-training quantization.
\newblock {\em CoRR}, abs/2302.05397, 2023.

\bibitem{DBLP:conf/cvpr/ParkAY17}
Eunhyeok Park, Junwhan Ahn, and Sungjoo Yoo.
\newblock Weighted-entropy-based quantization for deep neural networks.
\newblock In {\em 2017 {IEEE} Conference on Computer Vision and Pattern
  Recognition, {CVPR} 2017, Honolulu, HI, USA, July 21-26, 2017}, pages
  7197--7205. {IEEE} Computer Society, 2017.

\bibitem{DBLP:journals/corr/abs-2206-00820}
Sein Park, Junhyuk So, Juncheol Shin, and Eunhyeok Park.
\newblock {NIPQ:} noise injection pseudo quantization for automated {DNN}
  optimization.
\newblock {\em CoRR}, abs/2206.00820, 2022.

\bibitem{DBLP:conf/cvpr/RadosavovicKGHD20}
Ilija Radosavovic, Raj~Prateek Kosaraju, Ross~B. Girshick, Kaiming He, and
  Piotr Doll{\'{a}}r.
\newblock Designing network design spaces.
\newblock In {\em 2020 {IEEE/CVF} Conference on Computer Vision and Pattern
  Recognition, {CVPR} 2020, Seattle, WA, USA, June 13-19, 2020}, pages
  10425--10433. Computer Vision Foundation / {IEEE}, 2020.

\bibitem{DBLP:conf/cvpr/SandlerHZZC18}
Mark Sandler, Andrew~G. Howard, Menglong Zhu, Andrey Zhmoginov, and
  Liang{-}Chieh Chen.
\newblock Mobilenetv2: Inverted residuals and linear bottlenecks.
\newblock In {\em 2018 {IEEE} Conference on Computer Vision and Pattern
  Recognition, {CVPR} 2018, Salt Lake City, UT, USA, June 18-22, 2018}, pages
  4510--4520. Computer Vision Foundation / {IEEE} Computer Society, 2018.

\bibitem{DBLP:conf/iclr/StockFGGGJJ21}
Pierre Stock, Angela Fan, Benjamin Graham, Edouard Grave, R{\'{e}}mi Gribonval,
  Herv{\'{e}} J{\'{e}}gou, and Armand Joulin.
\newblock Training with quantization noise for extreme model compression.
\newblock In {\em 9th International Conference on Learning Representations,
  {ICLR} 2021, Virtual Event, Austria, May 3-7, 2021}. OpenReview.net, 2021.

\bibitem{DBLP:conf/cvpr/TanCPVSHL19}
Mingxing Tan, Bo Chen, Ruoming Pang, Vijay Vasudevan, Mark Sandler, Andrew
  Howard, and Quoc~V. Le.
\newblock Mnasnet: Platform-aware neural architecture search for mobile.
\newblock In {\em {IEEE} Conference on Computer Vision and Pattern Recognition,
  {CVPR} 2019, Long Beach, CA, USA, June 16-20, 2019}, pages 2820--2828.
  Computer Vision Foundation / {IEEE}, 2019.

\bibitem{DBLP:conf/cvpr/TungM18}
Frederick Tung and Greg Mori.
\newblock {CLIP-Q:} deep network compression learning by in-parallel
  pruning-quantization.
\newblock In {\em 2018 {IEEE} Conference on Computer Vision and Pattern
  Recognition, {CVPR} 2018, Salt Lake City, UT, USA, June 18-22, 2018}, pages
  7873--7882. Computer Vision Foundation / {IEEE} Computer Society, 2018.

\bibitem{DBLP:conf/iclr/UllrichMW17}
Karen Ullrich, Edward Meeds, and Max Welling.
\newblock Soft weight-sharing for neural network compression.
\newblock In {\em 5th International Conference on Learning Representations,
  {ICLR} 2017, Toulon, France, April 24-26, 2017, Conference Track
  Proceedings}. OpenReview.net, 2017.

\bibitem{wei2022qdrop}
Xiuying Wei, Ruihao Gong, Yuhang Li, Xianglong Liu, and Fengwei Yu.
\newblock Qdrop: Randomly dropping quantization for extremely low-bit
  post-training quantization.
\newblock {\em arXiv preprint arXiv:2203.05740}, 2022.

\bibitem{DBLP:journals/jstsp/WiedemannKMHMMN20}
Simon Wiedemann, Heiner Kirchhoffer, Stefan Matlage, Paul Haase, Arturo
  Marb{\'{a}}n, Talmaj Marinc, David Neumann, Tung Nguyen, Heiko Schwarz,
  Thomas Wiegand, Detlev Marpe, and Wojciech Samek.
\newblock Deepcabac: {A} universal compression algorithm for deep neural
  networks.
\newblock {\em {IEEE} J. Sel. Top. Signal Process.}, 14(4):700--714, 2020.

\bibitem{DBLP:journals/corr/abs-1905-08318}
Simon Wiedemann, Heiner Kirchhoffer, Stefan Matlage, Paul Haase, Arturo
  Marb{\'{a}}n, Talmaj Marinc, David Neumann, Ahmed Osman, Detlev Marpe, Heiko
  Schwarz, Thomas Wiegand, and Wojciech Samek.
\newblock Deepcabac: Context-adaptive binary arithmetic coding for deep neural
  network compression.
\newblock {\em CoRR}, abs/1905.08318, 2019.

\bibitem{DBLP:journals/corr/abs-2212-02770}
Lirui Xiao, Huanrui Yang, Zhen Dong, Kurt Keutzer, Li Du, and Shanghang Zhang.
\newblock {CSQ:} growing mixed-precision quantization scheme with bi-level
  continuous sparsification.
\newblock {\em CoRR}, abs/2212.02770, 2022.

\bibitem{DBLP:conf/icml/YaoDZGYTW0WMK21}
Zhewei Yao, Zhen Dong, Zhangcheng Zheng, Amir Gholami, Jiali Yu, Eric Tan,
  Leyuan Wang, Qijing Huang, Yida Wang, Michael~W. Mahoney, and Kurt Keutzer.
\newblock {HAWQ-V3:} dyadic neural network quantization.
\newblock In Marina Meila and Tong Zhang, editors, {\em Proceedings of the 38th
  International Conference on Machine Learning, {ICML} 2021, 18-24 July 2021,
  Virtual Event}, volume 139 of {\em Proceedings of Machine Learning Research},
  pages 11875--11886. {PMLR}, 2021.

\bibitem{DBLP:conf/iccv/YuM021}
Sixing Yu, Arya Mazaheri, and Ali Jannesari.
\newblock Auto graph encoder-decoder for neural network pruning.
\newblock In {\em 2021 {IEEE/CVF} International Conference on Computer Vision,
  {ICCV} 2021, Montreal, QC, Canada, October 10-17, 2021}, pages 6342--6352.
  {IEEE}, 2021.

\bibitem{DBLP:conf/icml/YuM022}
Sixing Yu, Arya Mazaheri, and Ali Jannesari.
\newblock Topology-aware network pruning using multi-stage graph embedding and
  reinforcement learning.
\newblock In Kamalika Chaudhuri, Stefanie Jegelka, Le Song, Csaba
  Szepesv{\'{a}}ri, Gang Niu, and Sivan Sabato, editors, {\em International
  Conference on Machine Learning, {ICML} 2022, 17-23 July 2022, Baltimore,
  Maryland, {USA}}, volume 162 of {\em Proceedings of Machine Learning
  Research}, pages 25656--25667. {PMLR}, 2022.

\bibitem{DBLP:conf/wacv/YuYGDKMK22}
Shixing Yu, Zhewei Yao, Amir Gholami, Zhen Dong, Sehoon Kim, Michael~W.
  Mahoney, and Kurt Keutzer.
\newblock Hessian-aware pruning and optimal neural implant.
\newblock In {\em {IEEE/CVF} Winter Conference on Applications of Computer
  Vision, {WACV} 2022, Waikoloa, HI, USA, January 3-8, 2022}, pages 3665--3676.
  {IEEE}, 2022.

\bibitem{DBLP:conf/eccv/ZhangYYH18}
Dongqing Zhang, Jiaolong Yang, Dongqiangzi Ye, and Gang Hua.
\newblock Lq-nets: Learned quantization for highly accurate and compact deep
  neural networks.
\newblock In Vittorio Ferrari, Martial Hebert, Cristian Sminchisescu, and Yair
  Weiss, editors, {\em Computer Vision - {ECCV} 2018 - 15th European
  Conference, Munich, Germany, September 8-14, 2018, Proceedings, Part {VIII}},
  volume 11212 of {\em Lecture Notes in Computer Science}, pages 373--390.
  Springer, 2018.

\bibitem{DBLP:conf/aaai/ZhouMCF18}
Yiren Zhou, Seyed{-}Mohsen Moosavi{-}Dezfooli, Ngai{-}Man Cheung, and Pascal
  Frossard.
\newblock Adaptive quantization for deep neural network.
\newblock In Sheila~A. McIlraith and Kilian~Q. Weinberger, editors, {\em
  Proceedings of the Thirty-Second {AAAI} Conference on Artificial
  Intelligence, (AAAI-18), the 30th innovative Applications of Artificial
  Intelligence (IAAI-18), and the 8th {AAAI} Symposium on Educational Advances
  in Artificial Intelligence (EAAI-18), New Orleans, Louisiana, USA, February
  2-7, 2018}, pages 4596--4604. {AAAI} Press, 2018.

\end{thebibliography}
